\documentclass{article}
\usepackage{spconf,amsmath,graphicx,float,array,subcaption,enumitem,array,tabularx,booktabs,mwe}

\usepackage{xcolor} 

\setlist{nosep, leftmargin=14pt}
\usepackage{amsmath} 
\usepackage{amssymb} 
\usepackage{bbold}
\usepackage{mathrsfs}

\def\Ours{{PWISeg}}

\title{PWISeg: Point-based Weakly-supervised Instance Segmentation for Surgical Instruments}
\name{Zhen Sun\textsuperscript{*}, Huan Xu\textsuperscript{*}, Jinlin Wu\textsuperscript{{\dag}}, Zhen Chen\textsuperscript{{\dag}}, Zhen Lei, Hongbin Liu\thanks{* Equal contribution.}\thanks{\textsuperscript{\dag}Corresponding Authors.}}
\address{Centre for Artificial Intelligence and Robotics, Hong Kong Institute of Science \& Innovation, \\Chinese Academy of Sciences, Hong Kong SAR, China}

\begin{document}
%
\maketitle
\begin{abstract}
 In surgical procedures, correct instrument counting is essential. Instance segmentation is a location method that locates not only an object's bounding box but also each pixel's specific details. However, obtaining mask-level annotations is labor-intensive in instance segmentation. To address this issue, we propose a novel yet effective weakly-supervised surgical instrument instance segmentation approach, named \textbf{P}oint-based \textbf{W}eakly-supervised \textbf{I}nstance \textbf{Seg}mentation (PWISeg). PWISeg adopts an FCN-based architecture with point-to-box and point-to-mask branches to model the relationships between feature points and bounding boxes, as well as feature points and segmentation masks on FPN, accomplishing instrument detection and segmentation jointly in a single model. Since mask level annotations are hard to available in the real world, for point-to-mask training, we introduce an unsupervised projection loss, utilizing the projected relation between predicted masks and bboxes as supervision signal. On the other hand, we annotate a few pixels as the key pixel for each instrument. Based on this, we further propose a key pixel association loss and a key pixel distribution loss, driving the point-to-mask branch to generate more accurate segmentation predictions. 
To comprehensively evaluate this task, we unveil a novel surgical instrument dataset with manual annotations, setting up a benchmark for further research. 
Our comprehensive research trial validated the superior performance of our {\Ours}. The results show that the accuracy of surgical instrument segmentation is improved, surpassing most methods of instance segmentation via weakly supervised bounding boxes. This improvement is consistently observed in our proposed dataset and when applied to the public HOSPI-Tools dataset.

\end{abstract}


\begin{figure}[h!]
  \centering
  \includegraphics[width=0.42\textwidth,keepaspectratio=false]{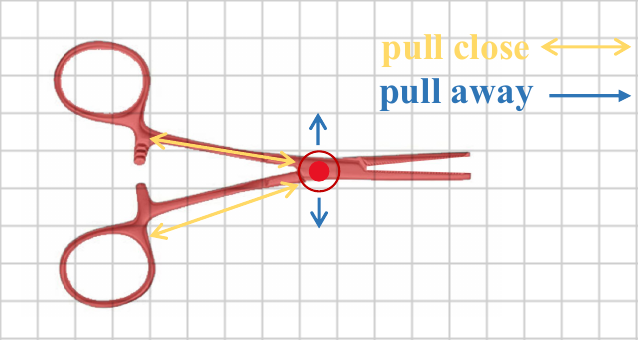}
  \caption{\textbf{Illustration of Anchor Based Possibility Loss.} Decrease the distance between anchor points and positive instances while increasing the distance between anchor points and negative instances}
  \label{fig:loss}
  \end{figure}

\section{Introduction}
\label{sec:intro}

In surgeries, it's very important to keep track of all the tools used. If any tool is left inside a person, it can cause infections or harm the body. Counting the tools accurately is a must. But, because there are so many different tools and surgeries can be prolonged and complicated, people can make mistakes when counting. This is why using computer technology to count the tools can be very helpful.

A significant challenge computer vision-based counting methods face in real operation room is the dense stacking and occlusion of instruments, making it difficult for existing detection methods to locate instruments accurately. To address this issue, instance segmentation offers a more precise localization technique, as it can locate instruments via joint bounding boxes and segmentation masks, improving the accuracy of instrument counting in occluded scenarios. Full supervision in instance segmentation requires resource-intensive mask-level annotations. In contrast, annotating bounding boxes and a few key points is more economical. Inspired by this, we propose a point-based weakly supervised method, PWISeg, for instance segmentation.

\begin{figure*}[h]
  \centering
  \includegraphics[width=\textwidth,keepaspectratio=true]{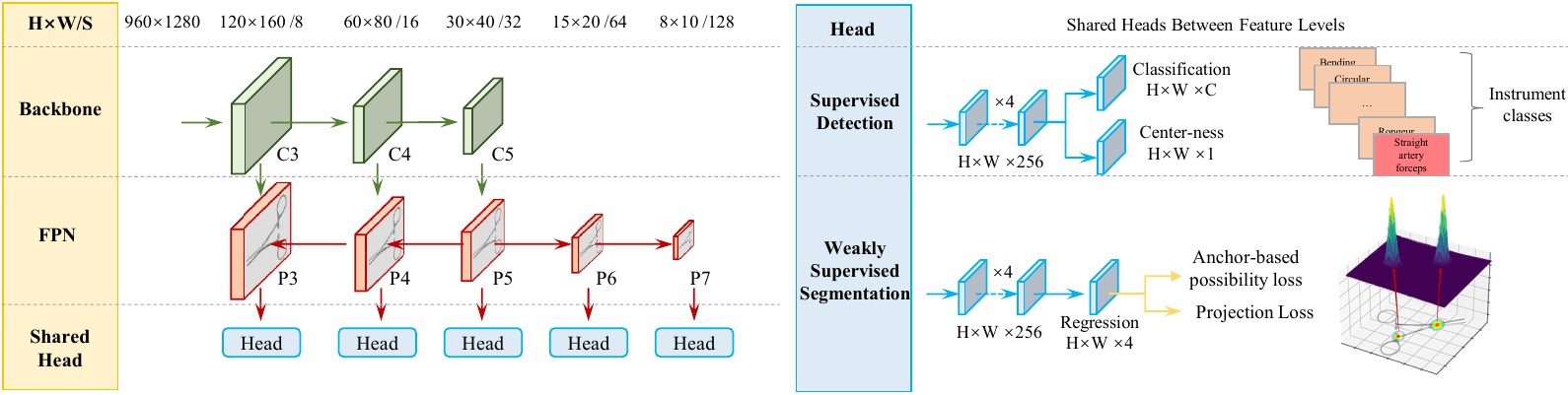}
  \caption{The overview of PWISeg framework. Our FCN-based model outputs a set of predicted bounding boxes and corresponding instance masks. Bounding boxes aim to determine the approximate location and size of objects, while instance masks provide detailed segmentation for each object.}
\label{fig:framework}
\end{figure*}

Our proposed weakly-supervised method, PWISeg, employs a FCN-based \cite{long2015fully} architecture that simulates the relationships between points and bounding boxes, as well as points and segmentation masks on the FPN \cite{lin2017feature}. This approach enables the simultaneous accomplishment of instrument detection and segmentation tasks within a single model. During the point-to-box training, we utilize Focal Loss \cite{lin2017focal} to assess the degree of congruence between the model’s predicted categories and the actual labels as a supervisory signal. In parallel, the Intersection over Union(IOU) Loss \cite{zhou2019iou} is used to evaluate the match between the model’s predicted bounding boxes and the actual ones. On the other hand, given the challenge of obtaining mask-level annotations in the real world, we introduce an unsupervised projection loss for point-to-mask training. This leverages the projection relationship between the predicted masks and bounding boxes as a supervisory signal. Furthermore, we annotate several key pixels on each instrument. Building on these, we propose a key pixel association loss and a key pixel distribution loss to drive the point-to-mask branch to generate more accurate segmentation predictions.

Additionally, we introduced a new surgical instrument dataset to alleviate the scarcity of professionally annotated data in this field. This dataset, which includes annotations of keypoints and bounding boxes, is expected to accelerate research and development in surgical instrument segmentation significantly. We achieved a mean Average Precision (mAP) on this dataset of $23.9\%$. We further validate the effectiveness of our PWISeg on the publicly available HOSPI-Tools dataset \cite{rodrigues2022evaluation}, with a mAP of $30.6\%$. It is worth noting that our method outperforms existing methods such as BoxInst \cite{tian2021boxinst}, Discobox \cite{lan2021discobox}, and BoxLevelSet \cite{li2022box} on both datasets.
\section{Methodology}
\label{sec:methodology}
Our PWISeg method is based on FCN architecture with point-to-box and point-to-mask branches to complete instance segmentation (Fig.~\ref{fig:framework}). For point-to-box branch, we use bounding boxes as supervision for training. For another, we use bounding boxes and a few key pixels as the weak supervision for training instead of mask labeling.

\subsection{Supervised Point to Box}
In box-level training, the network inputs an image of arbitrary size, which may contain multiple objects of interest. Each image within the labeled dataset explicitly indicates the category of the targets and their corresponding bounding boxes, annotated as ground-truth bounding boxes. These bounding boxes are defined as \({B_i}\), where \(B_i = \left(t_0^{(i)}, t_0^{(i)}, t_1^{(i)}, t_1^{(i)}, c^{(i)}\right) \in \mathbb{R}^4 \times \{1,2, \ldots, C\}\). Here, \(\left(t_0^{(i)}, y_0^{(i)}\right)\) corresponds to the top-left corner of the bounding box, \(\left(t_1^{(i)}, t_1^{(i)}\right)\) to the bottom-right corner, and \(c^{(i)}\) denotes the class of the object contained within the bounding box. In our dataset, there are 12 categories denoted by \(C\). The goal during the training process is to predict the target category and bounding boxes for each position in the images. Therefore, the loss function for box-level training is defined as follows:
\begin{equation}
\begin{aligned}
    L\left(\left\{\boldsymbol{c}_{(x,y)}\right\},\left\{\boldsymbol{t}_{(x,y)}\right\}\right) =\frac{1}{N_{\text{pos}}} \sum_{(x,y)} L_{\text{cls}}\left(\hat{c}_{(x,y)}, c_{(x,y)}\right) +\\\frac{\lambda}{N_{\text{pos}}} \sum_{(x,y)} \mathbb{1}_{\left\{c_{(x,y)}>0\right\}} L_{\text{reg}}\left(\hat{t_{(x,y)}}, \boldsymbol{t}_{(x,y)}\right)\text{ .}
\end{aligned}
\end{equation}
The set \( \left\{\hat{c}_{(x,y)}\right\} \) contains predicted class score vectors, whereas \( \left\{\hat{t}_{(x,y)}\right\} \) contains predicted target bounding boxes. \( N_{\text{pos}} \) denotes the number of positive samples, i.e., samples that contain a target. The loss function that balances classification accuracy and bounding box regression precision, optimizing the network's performance in detecting and categorizing each object. 

\subsection{Unsupervised Point To Mask}

\noindent \textbf{Unsupervised Projection Loss.} The projection-based loss function supervises the predicted mask against the annotated bounding boxes by employing the following definition:
\begin{equation}
\begin{aligned}
L_{\text{p}}
= Dice\left(\max _x(\hat{\boldsymbol{w}}), \max _x(\boldsymbol{t})\right) + \\
Dice\left(\max _y(\hat{\boldsymbol{w}}), \max _y(\boldsymbol{t})\right)\text{ ,}
\end{aligned}
\end{equation}
where \(\max _x(\hat{\boldsymbol{w}})\) represent the maximum values of the mask along the x-axis (y as y-axis), which effectively serve as the predicted boundaries and are somewhat analogous to the projection operations, while t stands for bounding box annotations. This loss function applies to all instances in the training image, with the final loss being their average.

\vspace{5px}
\noindent \textbf{Key-pixels association loss.} We leverage the affinity between pixels to diffuse the labels of a few key points within the entire bounding box to obtain mask pseudo-labels. Firstly, given a key pixel $I(i,j)$, the affinity between $(i,j)$ and corresponding nearest neighbor pixel $I(x,y)$ can be defined as : 
\begin{equation}
\begin{aligned}
A_{\{(i,j),(x,y)\}}=\hat{\boldsymbol{p}}_{i, j} \cdot \hat{\boldsymbol{p}}_{x, y}+\left(1-\hat{\boldsymbol{p}}_{i, j}\right) \cdot\left(1-\hat{\boldsymbol{p}}_{x, y}\right)\text{ ,}
\end{aligned}
\end{equation}
where $p$ is the probability that the pixel is a foreground pixel (object).
Then the pseudo-label for the associated pixel $(x,y)$ can be determined with a  threshold $\lambda$, as follow:
\begin{equation}
\begin{aligned}
\hat{y}_{(x,y)} = 
\begin{cases}
1, & \text{if } A_{\{(i, j),(x, y)\}} \geqslant \lambda\text{ ,} \\
0, & \text{otherwise}\text{ .}
\end{cases}
\end{aligned}
\end{equation}
When $\hat{y}_{(x,y)}=1$, it indicates that pixel $(x,y)$ has the same label as the key-pixel $(i,j)$ and belongs to the foreground; otherwise, $(x,y)$ is a background point. After determining the pseudo-labels around the key-pixels, we propagate labels outward based on intimacy until each pixel within the bounding box is assigned a pseudo-label. 

Finally, driven by the supervision of pseudo-labels, the pseudo mask loss can be defined as follows:
\begin{equation}
\begin{aligned}
\begin{array}{r}
L_{\text{ass}}=-\frac{1}{N} \sum_{(x,y)\in bbox} \hat{y}_{(x,y)} \log p_{(x,y)} 
+\\
(1-\hat{y}_{(x,y)}) \log p_{(x,y)}\text{ .}
\end{array}
\end{aligned}
\end{equation}
By focusing on the distribution of key pixels, we enhance the model's ability to recognize and associate relevant features within the bounding box.
\vspace{5px}

\noindent \textbf{Key-pixels distribution Loss.} 
When the output probability response is low, it is challenging for $A_{{(i,j),(x,y)}}$ to associate the key points with the surrounding points that belong to the foreground. To address this issue, we optimize the Wasserstein distance between the distribution of the key-pixels heatmap and the output probability as follows:

\begin{equation}\label{eq_heatmap}
\begin{aligned}
L_{\text{dis}}=\sum_{(x,y) \in {\rm bbox}}\|p_{(x,y)}-\Phi(h_{(x,y)})\|_{L_{1}}\text{ ,}
\end{aligned}
\end{equation}
where \( P \) represents the mask values predicted by the neural network, whereas \( \Phi (h(x, y)) \) denotes the heat map generated from key-pixels via the application of the Gaussian kernel function \( G(x, y) = \exp \left(-\frac{x^2 + y^2}{2\sigma^2}\right) \). This formula employs the L1 norm to quantify the disparity between the two heat maps within a designated bounding box. We use the probability distribution of key pixels' ground truth as supervision to provide a good starting point for the model.

In summary, we derive the anchor-based possibility loss as follows:
\begin{equation}
\begin{aligned}
L_{\text{seg}}= L_{\text{proj}} +\lambda_1  L_{\text{ass}}+\lambda_2 L_{\text{dis}}\text{ .}
\end{aligned}
\end{equation}
Therefore, applying this loss function in weakly supervised mask generation can significantly improve the model performance. By optimizing this loss function, we can more effectively guide the model to learn to extract key features from incompletely labeled data, leading to more accurate segmentation and mask generation.

\section{Dataset}
\label{sec:dataset}

In this work, we introduce a novel dataset designed for recognizing and categorizing surgical instruments. The high-resolution images ($1,280\times 960$ pixels) of the dataset detail the complexity of surgical instruments, particularly when grouped in settings akin to an operating room.

Our data acquisition strategy was multifaceted, ensuring a rich dataset that encapsulates the various scenarios under which surgical instruments are viewed. We employed a program-controlled camera mounted on the ceiling, which systematically rotated to photograph the instruments from multiple angles. Complementing this, we used a handheld camera to capture images of surgical instruments placed in trays on the floor, thus adding images with naturalistic shadows and lighting conditions to the dataset. Additionally, in a real-world operating room setting, we obtained close-up images of surgical instruments in use, capturing the tools in motion and from the surgical team's perspective.

Annotations in the dataset conform to the COCO \cite{lin2014microsoft} results format, supporting object and keypoint detection tasks. For object detection, we provide bounding boxes and class labels for each instrument. Keypoint annotations mark precise points on the instruments. Over 10,000 instruments have been annotated throughout the dataset. The dataset is divided into three sets for training, validation, and testing, containing $1788$, $200$ and $185$ images, respectively. 
\begin{figure}[t]
  \centering
  \includegraphics[width=0.45\textwidth]{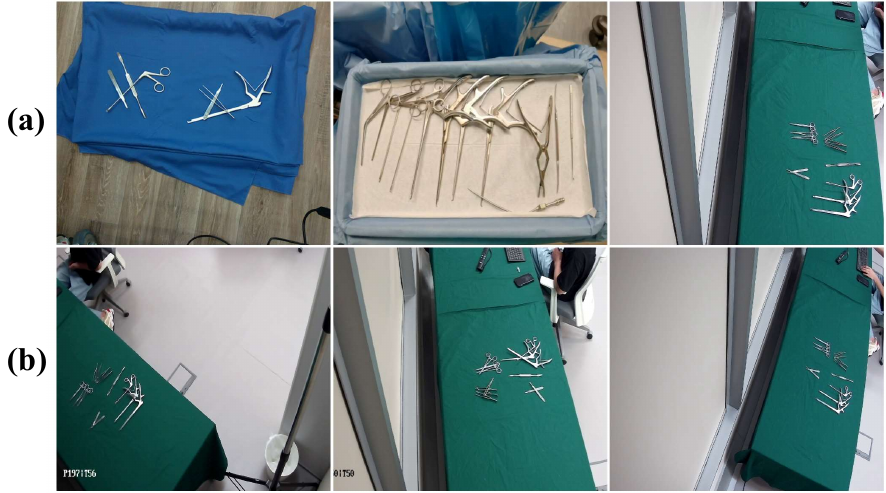}
  \caption{Different scenarios in our dataset.(a) is a multi-domain representation. (b) is a multi-view representation.}
  \label{fig:dataset}
\end{figure}

\begin{table*}[h]
\begin{tabularx}{\textwidth}{
   >{\centering\arraybackslash}p{2cm}
   >{\centering\arraybackslash}p{2cm}
   >{\centering\arraybackslash}p{1.6cm}
   >{\centering\arraybackslash}p{1.6cm}
   >{\centering\arraybackslash}p{1.6cm}
   >{\centering\arraybackslash}p{1.6cm}
   >{\centering\arraybackslash}p{1.6cm}
   >{\centering\arraybackslash}p{1.6cm}
}
    \toprule
    Method  & Backbone & \multicolumn{3}{c}{Detection} & \multicolumn{3}{c}{Segmentation}                                              \\
    \cmidrule(r){3-5} \cmidrule(r){6-8}
            &          & mAP                           & mAP$_{50}$                       & mAP$_{75}$ & mAP & mAP$_{50}$ & mAP$_{75}$ \\
    \midrule
    Discobox \cite{lan2021discobox} & ResNet-50   & $62.50$                             &  $90.40$                           & $73.40$           & $13.70$   & $36.40$   & $8.90$                   \\
    BoxLevelSet\cite{li2022box} & ResNet-50  & $87.90$                             & $71.20$                                & $63.60$          & $20.70$   & $69.40$   &$4.80$                   \\
    BoxInst \cite{tian2021boxinst} & ResNet-50   & $59.30$                             & $93.20$                                & $69.40$          & $21.30$   & $60.80$   &$13.00$                    \\
    Ours & ResNet-50   & $\boldsymbol{64.20}$                             & $\boldsymbol{96.80}$                                & $\boldsymbol{75.70 }$         & $\boldsymbol{23.90}$  & $\boldsymbol{66.30}$      &$\boldsymbol{13.80}$                 \\

    \bottomrule
\end{tabularx}
\caption{Performance metrics for object detection and segmentation on the dataset we proposed }
\label{tab:result_table1}
\end{table*}
\begin{table*}
\begin{tabularx}{\textwidth}{
   >{\centering\arraybackslash}p{2cm}
   >{\centering\arraybackslash}p{2cm}
   >{\centering\arraybackslash}p{1.6cm}
   >{\centering\arraybackslash}p{1.6cm}
   >{\centering\arraybackslash}p{1.6cm}
   >{\centering\arraybackslash}p{1.6cm}
   >{\centering\arraybackslash}p{1.6cm}
   >{\centering\arraybackslash}p{1.6cm}
}
    \toprule
    Method  & Backbone & \multicolumn{3}{c}{Detection} & \multicolumn{3}{c}{Segmentation}                                              \\
    \cmidrule(r){3-5} \cmidrule(r){6-8}
            &          & mAP                           & mAP$_{50}$                       & mAP$_{75}$ & mAP & mAP$_{50}$ & mAP$_{75}$ \\
    \midrule
    Discobox \cite{lan2021discobox} & ResNet-50   & $\boldsymbol{74.20 }$                            &$94.60$                             &$\boldsymbol{87.90}$            &$25.30$    & $74.10$   & $8.80$                    \\
    BoxLevelSet\cite{li2022box} & ResNet-50  &$72.60$                              &$94.30$                                 &$80.10$           &$28.10$    &$80.10$    & $10.30$                   \\
    BoxInst \cite{tian2021boxinst} & ResNet-50   & $66.00$                             &     $88.20$                            & $74.90$          &$29.10$    &  $77.10$  &    $15.00$                \\
    Ours & ResNet-50   & $73.20$                             & $\boldsymbol{95.20}$                                & $84.40$          & $\boldsymbol{30.60}$  & $\boldsymbol{80.50}$      &
    $\boldsymbol{15.80}$                 \\
    \bottomrule
\end{tabularx}
\caption{Performance metrics for object detection and segmentation on the HOSPI-Tools dataset 
}
\label{tab:result_table2}
\end{table*}
\setlength{\parskip}{0pt} 
\section{Experiments and results}
\label{sec:typestyle}

We assessed {\Ours} on the dataset we proposed and another public dataset. The experiment settings and results are reported in Section \ref{subsec:implementation_details} and Section \ref{subsec:main_results} respectively.

\subsection{Implementation Details}
\label{subsec:implementation_details}
We developed {\Ours} using the PyTorch framework and used the stochastic
gradient descent (SGD) algorithm \cite{robbins1951stochastic} to fine-tune {\Ours} on an Nvidia GeForce RTX $4090$ GPU. We started with a learning rate of $0.0001$ and use a batch size of $1$. The model was trained for $25,000$ iterations. We used two learning rate strategies. First, the LinearLR scheduler gradually increased the learning rate from the start until the $1000$th epoch. Then, the MultiStepLR scheduler reduced the learning rate by a factor of $10$ at the $17,000$th and $22,000$th iterations.
We will release the dataset and source code soon.

\subsection{Main Results}
\label{subsec:main_results}
\begin{figure}[tb]
\begin{minipage}[b]{.48\linewidth}
  \centering
  \centerline{\includegraphics[width=4.0cm,keepaspectratio=true]{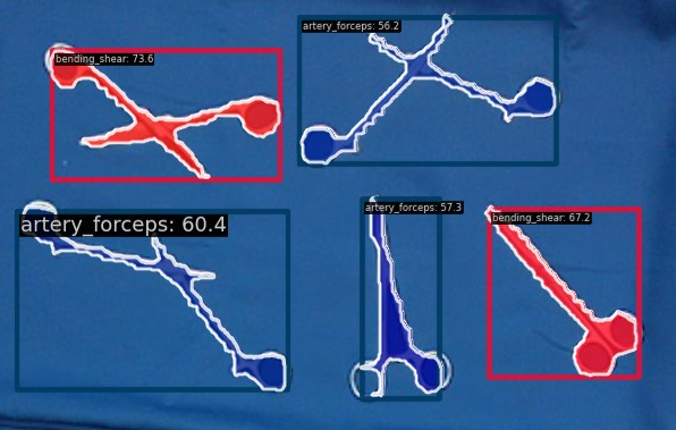}}
  \centerline{(a) Scattered scene}\medskip
\end{minipage}
\hfill
\begin{minipage}[b]{0.48\linewidth}
  \centering
  \centerline{\includegraphics[width=4.0cm,keepaspectratio=true]{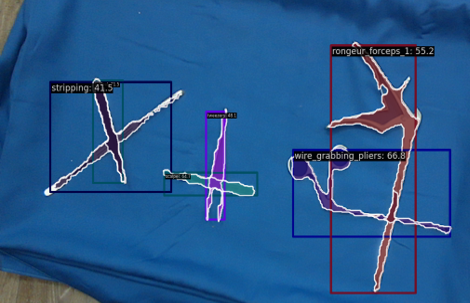}}
  \centerline{(b) Stacking scene}\medskip
\end{minipage}
\caption{Performance of {\Ours} in different scenarios, showcasing its ability to accurately segment and detect scattered surgical instruments and stacked surgical instruments.}
\label{fig:result}
\end{figure}
\noindent \textbf{Beyond Current Methods: Advancing Techniques.} Table {\ref{tab:result_table1}} shows that {\Ours}, using a ResNet-50 \cite{he2016deep} backbone, excelled in object detection and segmentation. It achieved high mAP scores (23.90 overall, 66.30 at 50$\%$ IoU, and 13.80 at 75$\%$ IoU) in segmentation and the highest scores in detection (64.20 overall, 96.80 at 50$\%$ IoU, and 75.70 at 75$\%$ IoU). These results are credited to the effective loss function used, enhancing the model's segmentation and detection accuracy.

\vspace{0.3cm}
\noindent \textbf{Proving Effectiveness: Testing on Public Datasets.} {\Ours} was tested on the HOSPI-Tools Dataset for surgical instruments. It demonstrated good adaptability and balance in detection and segmentation, as shown in Table {\ref{tab:result_table2}. In object detection, {\Ours} achieved a mAP of 73.20, slightly lower than Discobox's 74.20 but excelled with a 95.20 mAP at 50$\%$ overlap. In segmentation, it led with the highest mAP of 30.60, proving its effectiveness under this condition.

\section{Conclusion}
\label{sec:conclusion}

In this work, we introduces a novel dataset that is pivotal for the advancement of surgical instrument instance segmentation. By innovatively applying weakly supervised learning techniques to derive strong segmentation labels from bounding box annotations, and further refining segmentation accuracy through the strategic use of keypoints. We present an approach, {\Ours}, that significantly enhances the precision of instrument segmentation. This approach not only streamlines the annotation process but also promises substantial improvements in automated surgical tool recognition, with potential applications in enhancing real-time surgical assistance and operational efficiency within the medical field.

\newpage

\bibliographystyle{IEEEbib}
\bibliography{PWISeg}

\end{document}